\documentclass[10pt,twocolumn,letterpaper]{article}

\usepackage{cvpr}              

\usepackage{graphicx}
\usepackage{amsmath}
\usepackage{amssymb}
\usepackage{booktabs}
\usepackage{float}
\usepackage{caption}
\usepackage{colortbl}
\usepackage{xcolor}
\usepackage{float}
\usepackage[export]{adjustbox}
\usepackage[symbol]{footmisc}

\usepackage[pagebackref,breaklinks,colorlinks]{hyperref}

\usepackage[capitalize]{cleveref}
\crefname{section}{Sec.}{Secs.}
\Crefname{section}{Section}{Sections}
\Crefname{table}{Table}{Tables}
\crefname{table}{Tab.}{Tabs.}

\newcommand{\bB}{\mathbf{B}}

\newcommand{\bd}{\mathbf{d}}

\newcommand{\bF}{\mathbf{F}} 

\newcommand{\bI}{\mathbf{I}}

\newcommand{\bK}{\mathbf{K}}

\newcommand{\bR}{\mathbf{R}}

\newcommand{\bT}{\mathbf{T}}

\newcommand{\bx}{\mathbf{x}}\newcommand{\bX}{\mathbf{X}}

\newcommand{\btheta}{\boldsymbol{\theta}}

\newcommand{\nR}{\mathbb{R}}

\newcommand{\cL}{\mathcal{L}}

\newcommand{\figref}[1]{Fig.~\ref{#1}}

\newcommand{\eqnref}[1]{Eq.~\eqref{#1}}
\newcommand{\tabnref}[1]{Table~\ref{#1}}

\makeatletter
\DeclareRobustCommand\onedot{\futurelet\@let@token\@onedot}
\def\@onedot{\ifx\@let@token.\else.\null\fi\xspace}
\def\eg{e.g\onedot} 
\def\ie{i.e\onedot}

\def\etal{et~al\onedot}

\makeatother

\newcommand{\PAR}[1]{\vspace{0.1cm}\noindent{\bf #1} }

\newcommand{\norm}[1]{\left\lVert#1\right\rVert}

\begin{document}
	
\title{\includegraphics[scale=0.08]{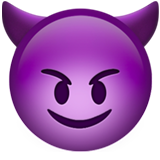}$\,$BAD-NeRF: Bundle Adjusted Deblur Neural Radiance Fields}

\author{Peng Wang$^{1,2}$\qquad Lingzhe Zhao$^{2}$\qquad Ruijie Ma$^{2}$\qquad Peidong Liu$^{2}$\footnotemark[2] \\$^{1}$Zhejiang University\qquad $^{2}$Westlake University
	\\{\tt\small \{wangpeng, zhaolingzhe, maruijie, liupeidong\}@westlake.edu.cn}}


\twocolumn[{
\maketitle
\vspace{-3.0em}
\begin{center}
	\setlength\tabcolsep{1pt}
	\begin{tabular}{ccccc}
		\includegraphics[width=0.195\textwidth]{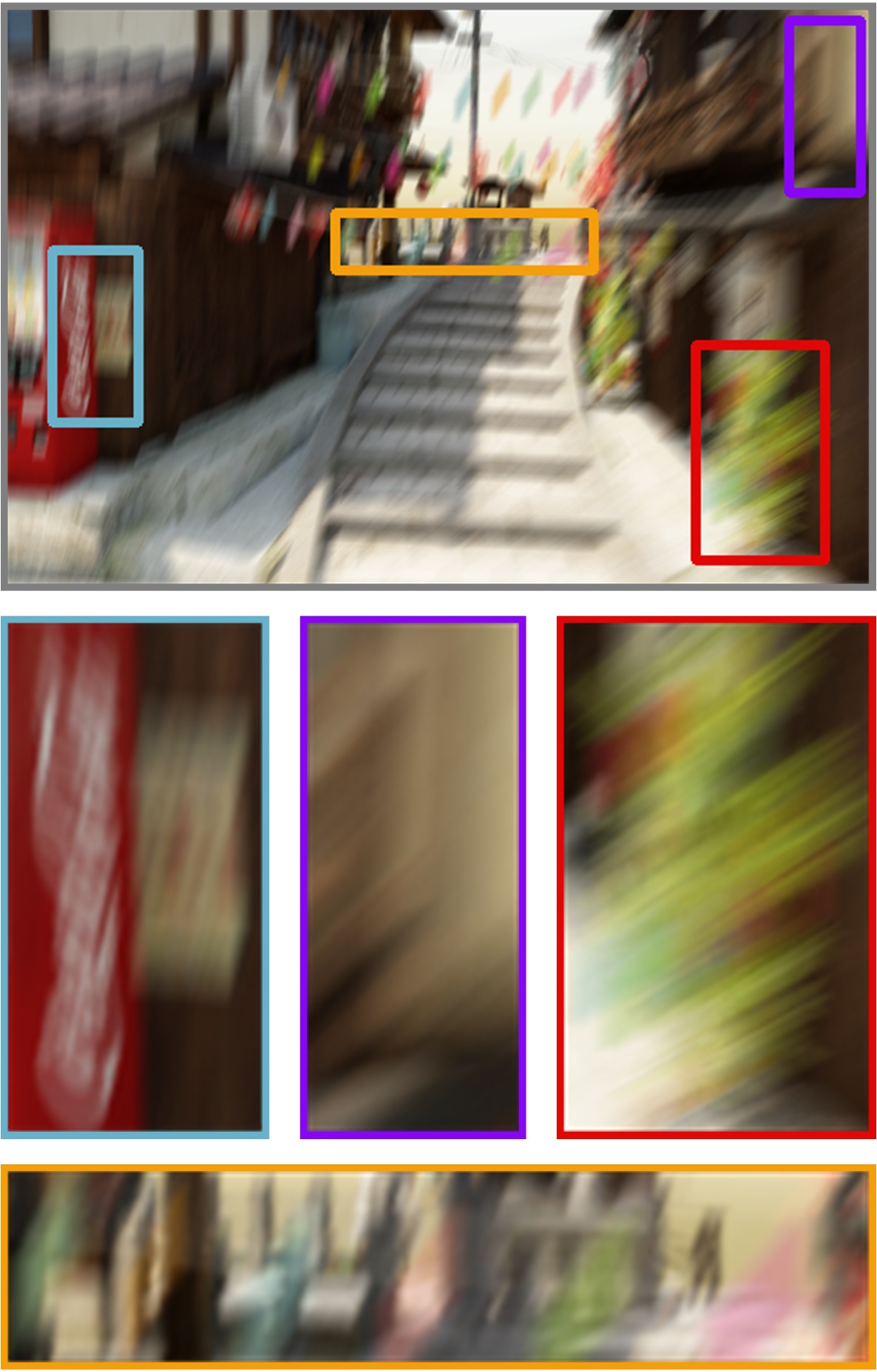} &
		\includegraphics[width=0.195\textwidth]{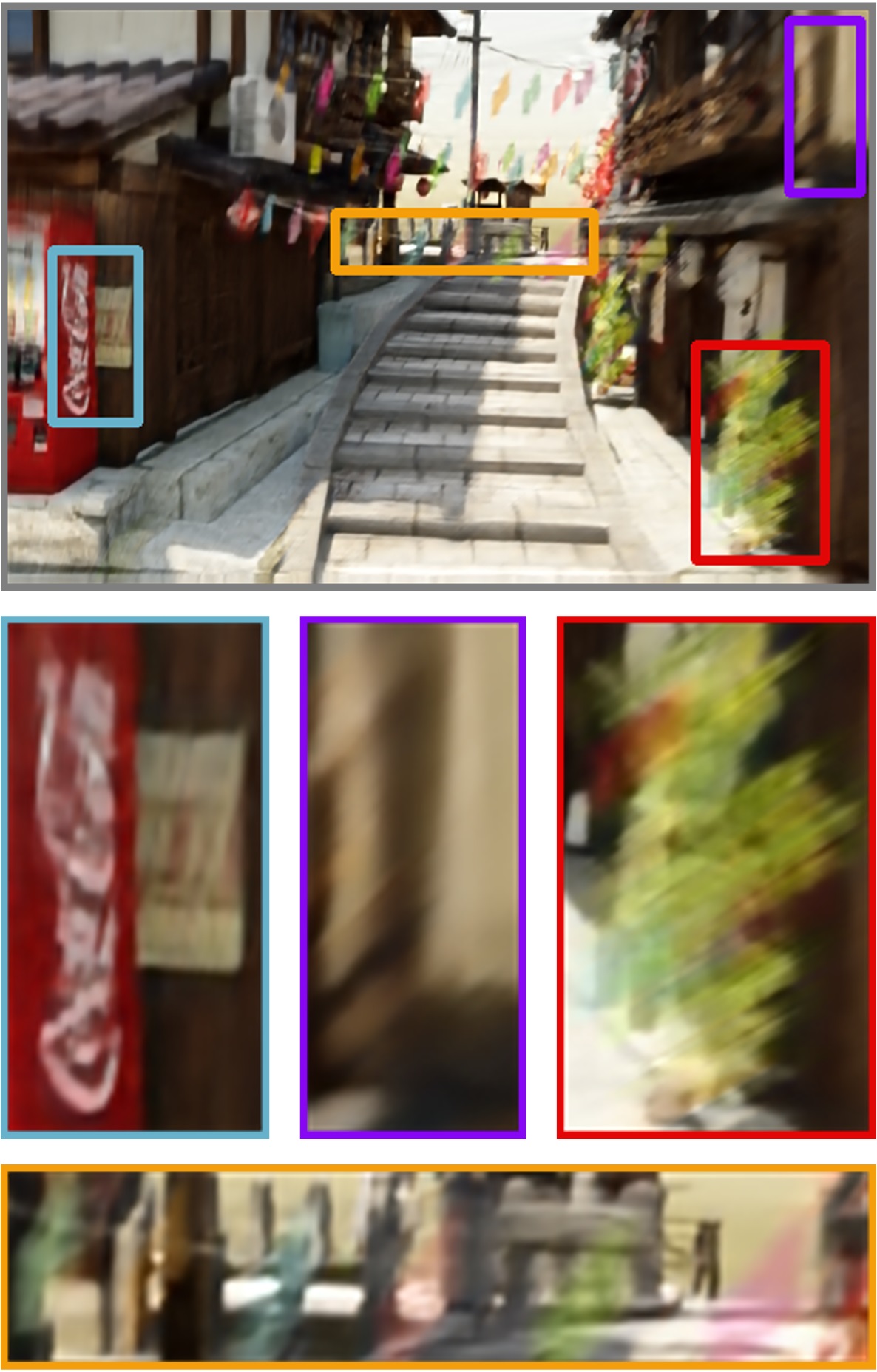} &
		\includegraphics[width=0.195\textwidth]{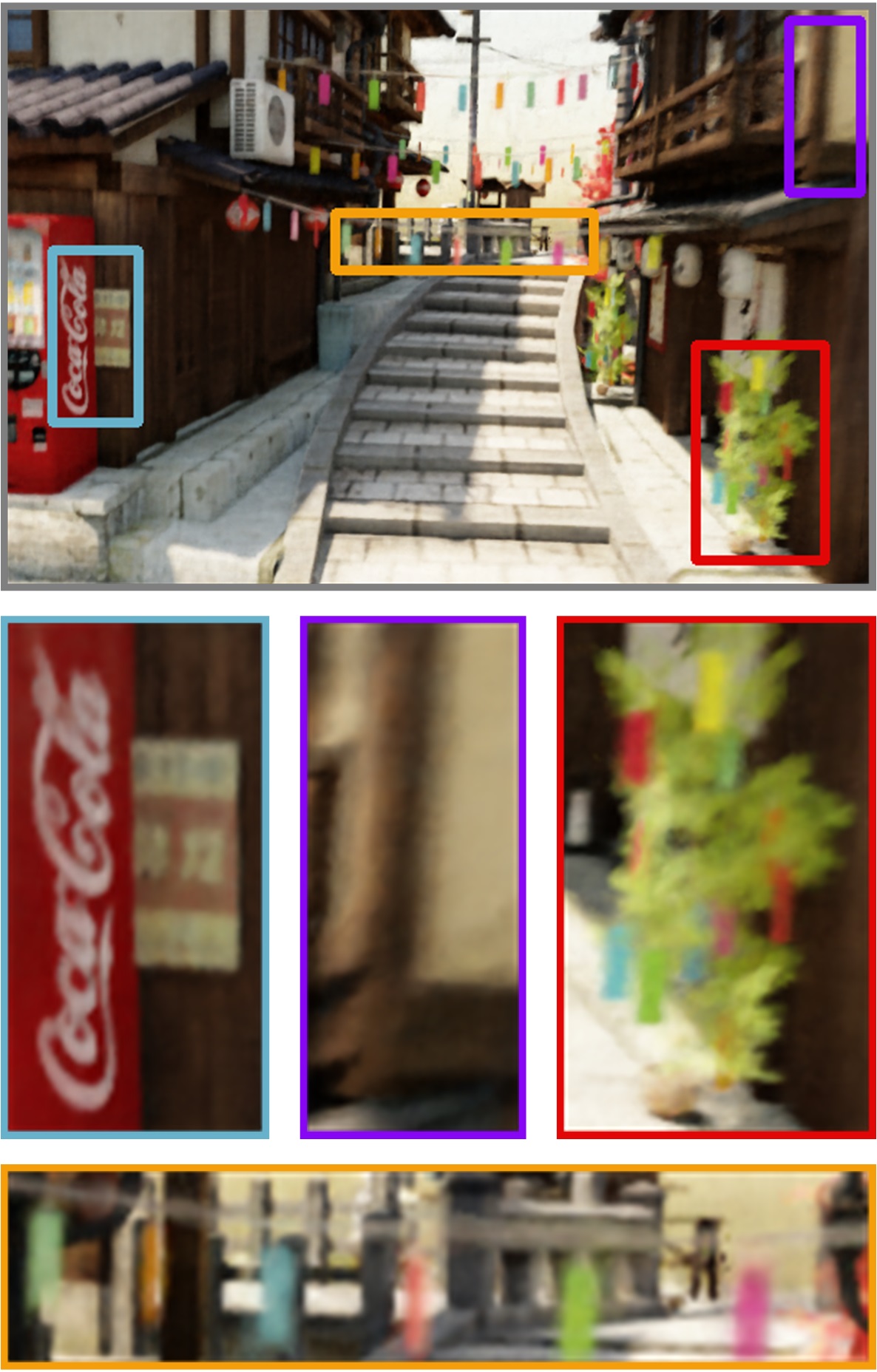} &
		\includegraphics[width=0.195\textwidth]{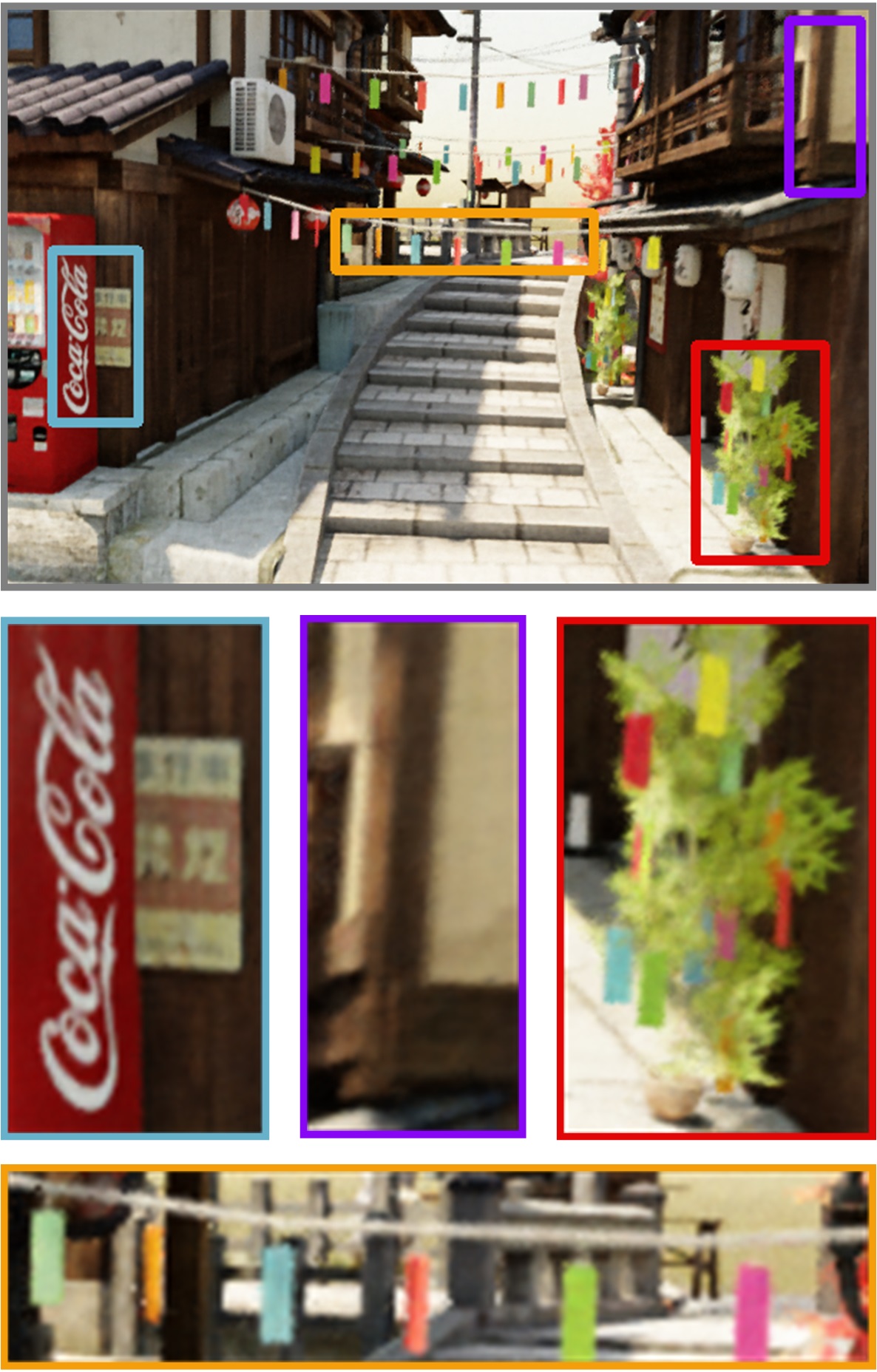} &
		\includegraphics[width=0.195\textwidth]{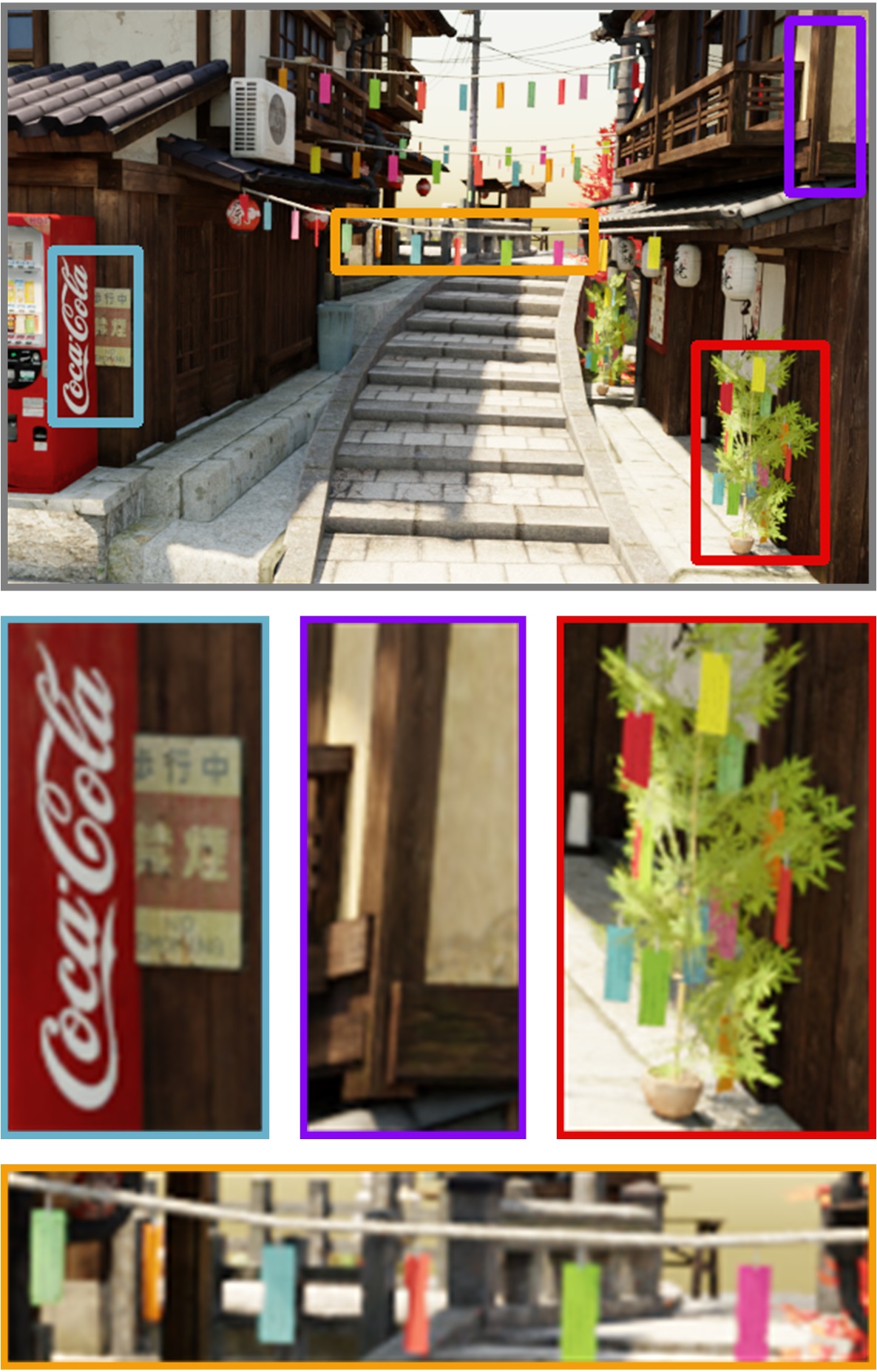} \\
		Input blurry image & SRN-Deblur \cite{Tao2018CVPR}  & Deblur-NeRF \cite{deblur-nerf} & BAD-NeRF & Ground truth
	\end{tabular}
	\vspace{-0.5em}
	\captionof{figure}{Given a set of severe motion blurred images, our bundle adjusted deblur NeRF (BAD-NeRF) jointly learns the neural radiance fields and recovers the camera motion trajectories within exposure time. It synthesizes novel images of higher quality than prior works.}
	\label{fig_teaser}
\end{center}
}]

\footnotetext[2]{Corresponding author.}

\begin{abstract}
\vspace{-0.6em}
Neural Radiance Fields (NeRF) have received considerable attention recently, due to its impressive capability in photo-realistic 3D reconstruction and novel view synthesis, given a set of posed camera images. Earlier work usually assumes the input images are of good quality. However, image degradation (e.g. image motion blur in low-light conditions) can easily happen in real-world scenarios, which would further affect the rendering quality of NeRF.
%
In this paper, we present a novel bundle adjusted deblur Neural Radiance Fields (BAD-NeRF), which can be robust to severe motion blurred images and inaccurate camera poses. Our approach models the physical image formation process of a motion blurred image, and jointly learns the parameters of NeRF and recovers the camera motion trajectories during exposure time. 
%
In experiments, we show that by directly modeling the real physical image formation process, BAD-NeRF achieves superior performance over prior works on both synthetic and real datasets. Code and data are available at \href{https://github.com/WU-CVGL/BAD-NeRF}{https://github.com/WU-CVGL/BAD-NeRF.}

\vspace{-1.5em}
\end{abstract}


\section{Introduction}
\label{sec:intro}
%
Acquiring accurate 3D scene geometry and appearance from a set of 2D images has been a long standing problem in computer vision. As a fundamental block for many vision applications, such as novel view image synthesis and robotic navigation, great progress has been made over the last decades. 
Classic approaches usually represent the 3D scene explicitly, in the form of 3D point cloud \cite{Furukawa2010PAMI,Whelan2015RSS}, triangular mesh \cite{Cornelis2008IJCV,Delaunoy2014CVPR,Gallup2010DAGM} or volumetric grid \cite{Niesner2013SIGGRAPH,Steinbruecker2014ICRA}. Recent advancements in implicit 3D representation by using a deep neural network, such as Neural Radiance Fields (NeRF) \cite{nerf}, have enabled photo-realistic 3D reconstruction and novel view image synthesis, given well posed multi-view images. 

NeRF takes a 5D vector (i.e. for spatial location and viewing direction of the sampled 3D point) as input and predicts its radiance and volume density via a multilayer perceptron. The corresponding pixel intensity or depth can then be computed by differentiable volume rendering\cite{volumerender, max1995optical}. 
%
While many methods have been proposed to further improve NeRF's performance, such as rendering efficiency \cite{Garbin2021, mueller2022instant}, training with inaccurate poses \cite{Lin2021} etc., limited work has been proposed to address the issue of training with motion blurred images. Motion blur is one of the most common artifacts that degrades images in practical application scenarios. It usually occurs in low-light conditions where longer exposure times are necessary. 
Motion blurred images would bring two main challenges to existing NeRF training pipeline: a) NeRF usually assumes the rendered image is sharp (i.e. infinitesimal exposure time), motion blurred image thus violates this assumption; b) accurate camera poses are usually required to train NeRF, however, it is difficult to obtain accurate poses from blurred images only, since each of them usually encodes information of the motion trajectory during exposure time. On the other hand, it is also challenging itself to recover accurate poses (e.g., via COLMAP \cite{colmap}) from a set of motion blurred images, due to the difficulties of detecting and matching salient keypoints. Combining both factors would thus further degrade NeRF's performance if it is trained with motion blurred images.
%

In order to address those challenges, we propose to integrate the real physical image formation process of a motion blurred image into the training of NeRF. We also use a linear motion model in the {\textbf{SE(3)}} space to represent the camera motion trajectory within exposure time. During the training stage, both the network weights of NeRF and the camera motion trajectories are estimated jointly.
%
In particular, we represent the motion trajectory of each image with both poses at the start and end of the exposure time respectively. The intermediate camera poses within exposure time can be linearly interpolated in the {\textbf{SE(3)}} space. This assumption holds in general since the exposure time is typically small. We can then follow the real physical image formation model of a motion blurred image to synthesize the blurry images. In particular, a sequence of sharp images along the motion trajectory within exposure time can be rendered from NeRF. The corresponding motion blurred image can then be synthesized by averaging those virtual sharp images. Both NeRF and the camera motion trajectories are estimated by minimizing the difference between the synthesized blurred images and the real blurred images. We refer this modified model as BAD-NeRF, i.e. bundle adjusted deblur NeRF. 

We evaluate BAD-NeRF with both synthetic and real datasets. The experimental results demonstrate that BAD-NeRF achieves superior performance compared to prior state of the art works (e.g. as shown in \figref{fig_teaser}), by explicitly modeling the image formation process of the motion blurred image.
%
In summary, our {\bf{contributions}} are as follows:
\vspace{-0.3em}
\begin{itemize}
	\itemsep0em 
	\item We present a photo-metric bundle adjustment formulation for motion blurred images under the framework of NeRF, which can be potentially integrated with other vision pipelines (e.g. a motion blur aware camera pose tracker \cite{mba-vo}) in future.
	\item We show how this formulation can be used to acquire high quality 3D scene representation from a set of motion blurred images. 
	\item We experimentally validate that our approach is able to deblur severe motion blurred images and synthesize high quality novel view images. 
\end{itemize}

\section{Related Work}
\label{sec:related}
We review two main areas of related works: neural radiance fields, and image deblurring. 

\PAR{Neural Radiance Field.} NeRF demonstrates impressive novel view image synthesis performance and 3D scene representation capability \cite{nerf}. Many variants of NeRF have been proposed recently. For example, \cite{Wizadwongsa2021,Lassner2021,Rebain2021,Yu2021,Reiser2021,Garbin2021,Piala2022} proposed methods to improve the rendering efficiency of NeRF, such that it can render images in real-time. \cite{MartinBrualla2021,Mildenhall2022,Lin2021,Jeong2021,deblur-nerf} proposed to improve the training performance of NeRF with inaccurate posed images or images captured under challenging conditions. There are also methods that extend NeRF for large scale scene representations \cite{Xiangli2021,Tancik2022,Turki2022} and non-rigid object reconstructions (e.g. human body) \cite{Reiser2021,Park2020,Pumarola2021,Peng2021,Gafni2021,Peng2021a,Weng2022,Athar2022}. NeRF is also recently being used for 3D aware image generation models \cite{Niemeyer2021,Wang2022,Liu2021,OrEl2021,Deng2022,Gu2022}.

We will mainly detailed review those methods which are the closest to our work in this section. 
BARF proposed to optimize camera poses of input images as additional variables together with parameters of NeRF \cite{Lin2021}. They propose to gradually apply the positional encoding to better train the network as well as the camera poses. A concurrent work from Jeong et al. \cite{Jeong2021} also proposed to learn the scene representation and camera parameters (i.e. both extrinsic and intrinsic parameters) jointly.
Different from BARF \cite{Lin2021} and the work from \cite{Jeong2021}, which estimate the camera pose at a particular timestamp, our work proposes to optimize the camera motion trajectory within exposure time. 
Deblur-NeRF aims to train NeRF from a set of motion blurred images \cite{deblur-nerf}. They obtain the camera poses from either ground truth or COLMAP \cite{colmap}, and fix them during training stage. Severe motion blurred images challenge the pose estimations (e.g. from COLMAP), which would thus further degrade the training performance of NeRF.
Instead of relying heavily on the accurately posed images, our method estimates camera motion trajectories together with NeRF parameters jointly. The resulting pipeline is thus robust to inaccurate initialized camera poses due to severe motion blur. 

\PAR{Image Deblurring.} Existing techniques to solve motion deblurring problem can be generally classified into two main categories: the first type of approach formulates the problem as an optimization problem, where the latent sharp image and the blur kernel are optimized using gradient descent during inference \cite{Cho2009ToG,Fergus2006SIGGGRAPH,Krishnan2009NIPS,Levin2009CVPR,Shan2008ToG,Xu2010ECCV,park2017joint}. Another type of approaches phrases the task as an end-to-end learning problem. Building upon the recent advances of deep convolution neural networks, state-of-the-art results have been obtained for both single image deblurring \cite{Nah2017CVPR,Tao2018CVPR,Kupyn2019ICCV} and video deblurring \cite{Su2017CVPR}. 

The closest work to ours is from Park et al. \cite{park2017joint}, which jointly recovers the camera poses, dense depth maps, and latent sharp images from a set of multi-view motion blurred images. They formulate the problem under the classic optimization framework, and aim to maximize both the self-view photo-consistency and cross-view photo-consistencies. Instead of representing the 3D scene as a set of multi-view dense depth maps and latent sharp images, we represent it with NeRF implicitly, which can better preserve the multi-view consistency and enable novel view image synthesis.
\section{Method}\label{sec:method}
\begin{figure*}
	\centering
	\includegraphics[width=0.9\linewidth]{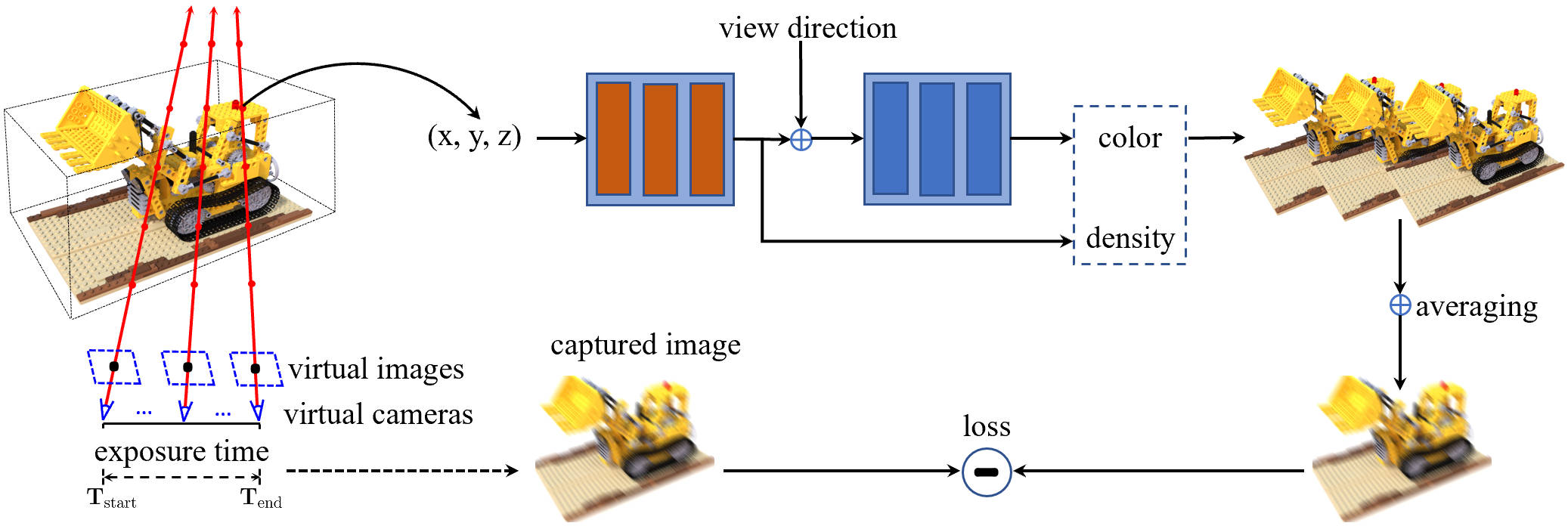}
	\caption{{\bf{The pipeline of BAD-NeRF.}} Given a set of motion blurred images, we train NeRF to learn the 3D scene representation. Different from prior works, which usually model the camera pose at a fixed timestamp, we represent the motion trajectory of each image with both poses at the start and end of the exposure time respectively. Intermediate virtual camera poses can then be linearly interpolated in {\textbf{SE3}} space. We then follow the standard neural rendering procedures of NeRF to synthesize those virtual sharp images within exposure time. The blurry image can then be synthesized by averaging those virtual images, which obeys the real physical image formation process of a motion blurred image. The whole network and both the start and end poses are jointly estimated by minimizing the photo-metric loss between the synthesized and real blurry images.}
	\label{fig_overview}
	\vspace{-1.5em}
\end{figure*}
In this section, we present the details of our bundle adjusted deblur NeRF (BAD-NeRF). BAD-NeRF learns both the 3D scene representation and recovers the camera motion trajectories, given a set of motion blurred images as shown in \figref{fig_overview}. We follow the real physical image formation process of a motion blurred image to synthesize blurry images from NeRF. Both NeRF and the motion trajectories are estimated by maximizing the photo-metric consistency between the synthesized blurry images and the real blurry images. We will detail each component as follows. 

\subsection{Neural Radiance Fields}
We follow the general architecture of NeRF \cite{nerf} to represent the 3D scene implicitly using two Multi-layer Perceptrons (MLP). To query the pixel intensity $\bI(\bx)$ at pixel location $\bx$ for a particular image with pose $\bT_c^w$, we can shoot a ray into the 3D space. The ray is defined by connecting the camera center and the corresponding pixel. We can then compute the pixel intensity by volume rendering along the ray \cite{max1995optical}. This process can be formally described as follows. 

Given a particular 3D point $\bX^w$ with depth $\lambda$ along the ray, we can obtain its coordinate defined in the world coordinate frame as:
\begin{align}
	\bd^c &= \bK^{-1} \begin{bmatrix}
		\bx \\
		1
	\end{bmatrix}, \\
	\bX^w &= \bT_c^w \cdot \lambda \bd^c, 
\end{align}
where $\bx$ is the pixel location, $\bK$ is the camera intrinsic parameters by assuming a simple pinhole camera model, $\bd^c$ is the ray direction defined in the camera coordinate frame, $\bT_c^w$ is the camera pose defined from camera frame to world frame. We can then query NeRF for its corresponding view-depend color $\boldsymbol{c}$ and volume density $\sigma$ using MLPs:
\begin{equation}
	(\boldsymbol{c}, \sigma) = \bF_\theta(\bX^w, \quad \bR_c^w \cdot \bd^c),
\end{equation}
where $\bF_\theta$ is the Multi-layer Perceptrons parameterized with learnable parameters $\btheta$, $\bR_c^w$ is the rotation matrix which transforms the viewing direction $\bd^c$ from camera coordinate frame to world coordinate frame. Following Mildenhall et al. \cite{nerf}, which showed that coordinate-based approaches struggle with learning details from low-dimensional inputs, we also use their proposed Fourier embedding $\gamma(\bX)$ representation of a 3D point $\bX$ and $\gamma(\bd)$ representation of the viewing direction $\bd$, to map the low-dimensional inputs to high-dimensional space. 

Following the definition of volume rendering \cite{max1995optical}, we can then compute the pixel intensity by sampling 3D points along the ray as follows:
\begin{equation}
	\bI(\bx) = \sum_{i=1}^{n} T_i(1-{\rm exp}(-\sigma_i\delta_i))\boldsymbol{c}_i,
\end{equation}
where $n$ is the number of sampled 3D points along the ray, both $\boldsymbol{c}_i$ and $\sigma_i$ are the predicted color and volume density of the $i^{th}$ sampled 3D point via $\bF_\theta$, $\delta_i$ is the distance between the $i^{th}$ and $(i+1)^{th}$ sampled point, $T_i$ is the transmittance factor which represents the probability that the ray does not hit any particle until the $i^{th}$ sampled point. $T_i$ can be formally defined as:
\begin{equation}
	T_i = {\textrm{exp}} (-\sum_{k=1}^{i-1}\sigma_k\delta_k),
\end{equation}
where $\sigma_k$ is the predicted volume density for $k^{th}$ point by $\bF_\theta$, $\delta_k$ is the corresponding distance between neighboring points.

The above derivations show that the rendered pixel intensity $\bI(\bx)$ is a function of the MLPs with learnable parameters $\btheta$, as well as the corresponding camera pose $\bT_c^w$. It can also be derived that $\bI(\bx)$ is differentiable with respect to both $\btheta$ and $\bT_c^w$, which lays the foundations for our bundle adjustment formulation with a set of motion blurred images.

\subsection{Motion Blur Image Formation Model}
The physical image formation process refers to a digital camera collecting photons during the exposure time and converting them into measurable electric charges. The mathematical modeling of this process involves integrating over a set of virtual sharp images: 
\begin{equation}
	\bB(\bx) = \phi \int_{0}^{\tau} \bI_\mathrm{t}(\bx) \mathrm{dt},
\end{equation}
where $\bB(\bx) \in \nR^{\mathrm{W} \times \mathrm{H} \times 3}$ is the captured image, $\mathrm{W}$ and $\mathrm{H}$ are the width and height of the image respectively, $\bx \in \nR^2$ represents the pixel location, $\phi$ is a normalization factor, $\tau$ is the camera exposure time, $\bI_\mathrm{t}(\bx) \in \nR^{\mathrm{W} \times \mathrm{H} \times 3}$ is the virtual sharp image captured at timestamp $\mathrm{t}$ within the exposure time. A blurred image $\bB(\bx)$ caused by camera motion during the exposure time,  is formed by different virtual images $\bI_\mathrm{t}(\bx)$ for each $t$. The model can be discretely approximated as 
\begin{equation}\label{eq_blur_im_formation}
	\bB(\bx)  \approx \frac{1}{n} \sum_{i=0}^{n-1} \bI_\mathrm{i}(\bx), 
\end{equation}
where $n$ is the number of discrete samples. 

The degree of motion blur in an image thus depends on the camera motion during the exposure time. For example, a fast-moving camera causes little relative motion for shorter exposure time, whereas a slow-moving camera leads to a motion blurred image for long exposure time (e.g.~in low light conditions). It can be further derived that $\bB(\bx)$ is differentiable with respect to each of virtual sharp images $\bI_i(\bx)$.

\subsection{Camera Motion Trajectory Modeling}
As derived in \eqnref{eq_blur_im_formation}, we need to model the corresponding poses of each latent sharp image within exposure time, so that we can render them from NeRF (i.e. $\bF_\theta$). We approximate the camera motion with a linear model during exposure time which is usually small (e.g. $\le$ 200 ms). Specifically, two camera poses are parameterized, one at the beginning of the exposure $\bT_\mathrm{start} \in \mathbf{SE}(3)$ and one at the end $\bT_\mathrm{end} \in \mathbf{SE}(3)$. Between these two poses, we linearly interpolate poses in the Lie-algebra of $\mathbf{SE}(3)$. The virtual camera pose at time $t \in [0,\tau]$ can thus be represented as 
\begin{equation} \label{eq_trajectory}
	\bT_t = \bT_\mathrm{start} \cdot \mathrm{exp}(\frac{t}{\tau} \cdot \mathrm{log}(\bT_\mathrm{start}^{-1} \cdot \bT_\mathrm{end})),
\end{equation} 
where $\tau$ is the exposure time. $\frac{t}{\tau}$ can be further derived as $\frac{i}{n-1}$ for the $i^{th}$ sampled virtual sharp image (i.e. with pose as $\bT_i$), when there are $n$ images being sampled in total. It can be derived that $\bT_i$ is differentiable with respect to both $\bT_{\mathrm{start}}$ and $\bT_{\mathrm{end}}$. For more details on the interpolation and derivations of the related Jacobian, please refer to prior work from Liu et al. \cite{mba-vo}. The goal of BAD-NeRF is now to estimate both $\bT_\mathrm{start}$ and $\bT_\mathrm{end}$ for each frame, as well as the learnable parameters of $\bF_\theta$. 

Besides the linear interpolation approach, we also explore the trajectory representation with a higher order spline (\ie cubic B-Spline), which can represent more complex camera motions. Since the exposure time is usually relatively short, we find that a linear interpolation can already deliver satisfying performance from the experimental results. More details on the cubic B-Spline formulation can be found in our supplementary material. 

\subsection{Loss Function}
Given a set of $K$ motion blurred images, we can then estimate the learnable parameters $\btheta$ of NeRF as well as the camera motion trajectories for each image (i.e. $\bT_\mathrm{start}$ and $\bT_\mathrm{end}$) by minimizing the photo-metric loss:
\begin{equation}
	\cL = \sum_{k=0}^{K-1} \norm{\bB_k(\bx) - \bB^{gt}_k(\bx)},
\end{equation}
where $\bB_k(\bx)$ is the $k^{th}$ blurry image synthesized from NeRF by following the above image formation model, $\bB^{gt}_k(\bx)$ is the corresponding real captured blurry image. 

To optimize the learnable parameter $\btheta$, $\bT_\mathrm{start}$ and $\bT_\mathrm{end}$ for each image, we need to have the corresponding Jacobians:
\begin{equation}
	\frac{\partial \cL}{\partial \btheta} = \sum_{k=0}^{K-1} \frac{\partial \cL}{\partial \bB_k(\bx)} 
	\cdot \frac{1}{n}\sum_{i=0}^{n-1} \frac{\partial \bB_k(\bx)}{\partial \bI_i(\bx)} 
	\frac{\partial \bI_i(\bx)}{\partial \btheta},
\end{equation}
\begin{equation}
	\frac{\partial \cL}{\partial \bT_\mathrm{start}} = \sum_{k=0}^{K-1} \frac{\partial \cL}{\partial \bB_k(\bx)} 
	\cdot \frac{1}{n}\sum_{i=0}^{n-1} \frac{\partial \bB_k(\bx)}{\partial \bI_i(\bx)} 
	\frac{\partial \bI_i(\bx)}{\partial \bT_\mathrm{start}},
\end{equation}
\begin{equation}
	\frac{\partial \cL}{\partial \bT_\mathrm{end}} = \sum_{k=0}^{K-1} \frac{\partial \cL}{\partial \bB_k(\bx)} 
	\cdot \frac{1}{n}\sum_{i=0}^{n-1} \frac{\partial \bB_k(\bx)}{\partial \bI_i(\bx)} 
	\frac{\partial \bI_i(\bx)}{\partial \bT_\mathrm{end}}.
\end{equation}
We parameterize both $\bT_\mathrm{start}$ and $\bT_\mathrm{end}$ with their corresponding Lie algebras of $\mathbf{SE}(3)$, which can be represented by a 6D vector respectively. 

\section{Experiments}
\label{sec:experiment}
\subsection{Experimental details}
\PAR{Benchmark datasets.} To evaluate the performance of our network, we use both the synthetic datasets and real datasets from prior works, \ie the datasets from Deblur-NeRF \cite{deblur-nerf} and MBA-VO \cite{mba-vo}. The synthetic dataset from Deblur-NeRF \cite{deblur-nerf} is synthesized by using Blender \cite{Blender}. The datasets are generated from 5 virtual scenes, assuming the camera motion is in constant velocity within exposure time. Both the anchor camera poses and camera velocities (\ie in 6 DoFs) within exposure time are randomly sampled, \ie they are not sampled along a continuous motion trajectory. The blurry image is then generated by averaging the sampled virtual images within the exposure time for each camera. To synthesize more realistic blurry images, we increase the number of virtual images to 51 from 10 and keep the other settings fixed. They also capture a real blurry dataset by deliberately shaking a handheld camera. 

To investigate the performance of our method more thoroughly, we also evaluate our method on a dataset for motion blur aware visual odometry benchmark (\ie MBA-VO \cite{mba-vo}). The dataset contains both synthetic and real blurry images. The synthetic images are synthesized from an Unreal game engine and the real images are captured by a handheld camera within indoor environment. Different from the synthetic dataset from Deblur-NeRF \cite{deblur-nerf}, the synthetic images from MBA-VO \cite{mba-vo} are generated based on real motion trajectories (\ie not constant velocity) from the ETH3D dataset \cite{schops2019bad}.

\PAR{Baseline methods and evaluation metrics.} We evaluate our method against several state-of-the-art learning-based deblurring methods, \ie SRNDeblurNet\cite{Tao2018CVPR}, PVD\cite{son2021recurrent}, MPR\cite{zamir2021multi}, Deblur-NeRF\cite{deblur-nerf} as well as a classic multi-view image deblurring method from Park \etal \cite{park2017joint}. For deblurring evaluation, we synthesize images corresponding to the middle virtual camera poses (\ie the one at the middle of the exposure time) from the trained NeRF, and evaluate its performance against the other methods. 
Since SRNDeblurNet\cite{Tao2018CVPR}, PVD\cite{son2021recurrent} and  MPR\cite{zamir2021multi} are primarily designed for single image deblurring, they are not generically suitable for novel view image synthesis. We thus firstly apply them to each blurry image to get the restored images, and then input those restored images to NeRF \cite{nerf} for training. For novel view image synthesis evaluation, we render novel-view images with their corresponding poses from those trained NeRF. All the methods use the estimated poses from COLMAP \cite{colmap} for NeRF training. To investigate the sensitivity of Deblur-NeRF \cite{deblur-nerf} to the accuracy of the camera poses, we also train it with the ground-truth poses provided by Blender during dataset generation. 

The quality of the rendered image is evaluated with the commonly used metrics, \ie the PSNR, SSIM and LPIPS\cite{zhang2018unreasonable} metrics.
We also evaluate the absolute trajectory error (ATE) against the method from Park \etal \cite{park2017joint} and BARF \cite{Lin2021} on the evaluation of the estimated camera poses. The ATE metric is commonly used for visual odometry evaluations \cite{mba-vo}.
\begin{figure}
	\includegraphics[width=0.9\columnwidth]{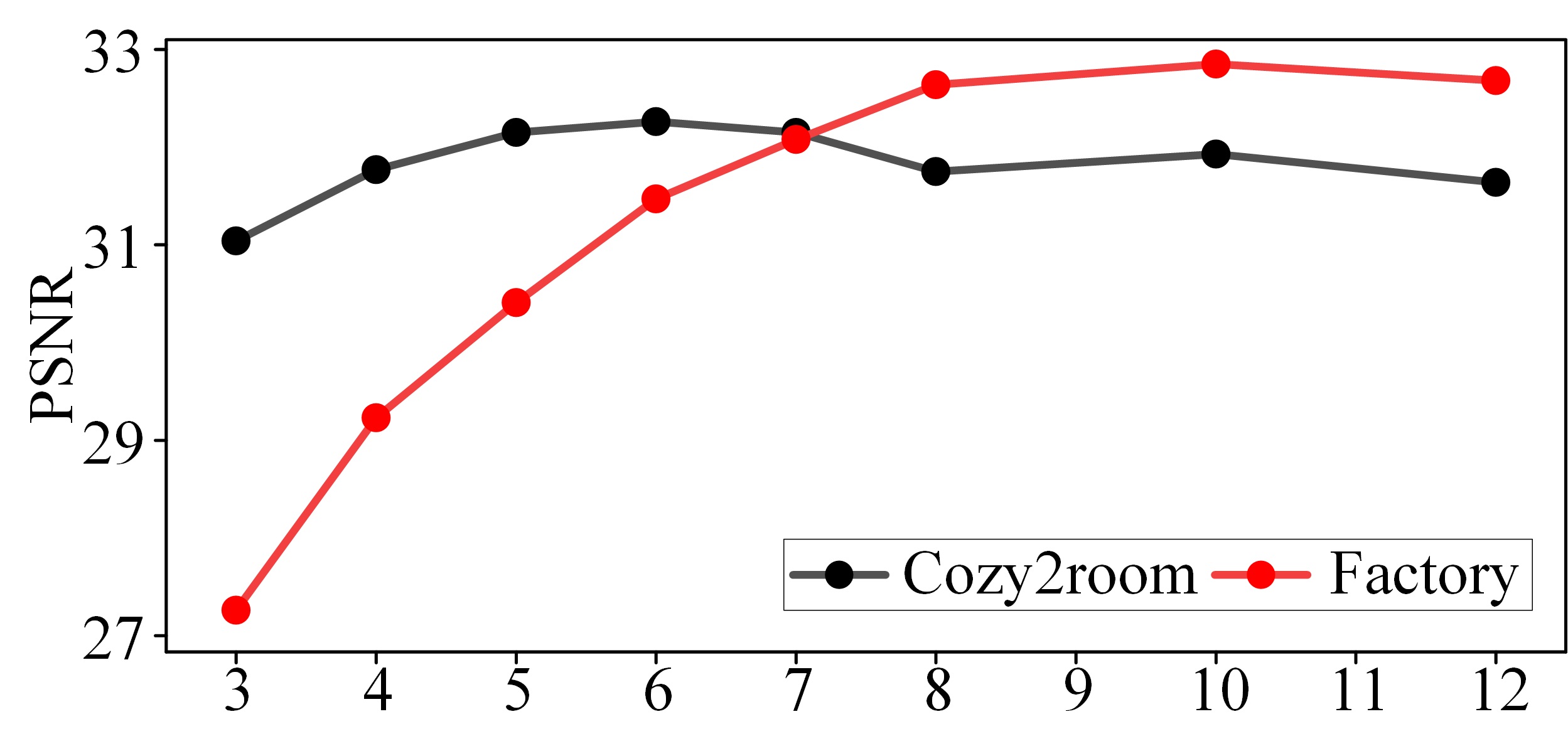}
	\vspace{-0.8em}
	\caption{\textbf{The effect of the number of interpolated virtual cameras.} The results demonstrate that the performance saturates as the number increases.}
	\label{fig_ablation}
	\vspace{-0.4em}
\end{figure}
\begin{table}[t]
	\vspace{-0.7em}
	\setlength\tabcolsep{1pt}
	\resizebox{\linewidth}{!}{
		\begin{tabular}{c|ccc|ccc}
			\specialrule{0.1em}{1pt}{1pt}
			& \multicolumn{3}{c|}{\textit{Deblur-NeRF}} & \multicolumn{3}{c}{\textit{MBA-VO}}\\
			& PSNR$\uparrow$ & SSIM$\uparrow$ & LPIPS$\downarrow$ & PSNR$\uparrow$ & SSIM$\uparrow$ & LPIPS$\downarrow$ \\
			\specialrule{0.05em}{1pt}{1pt}
			Direct optimization & 29.99 & .8737 & .0996 & 28.86 & .8454 & .2044 \\
			Cubic B-Spline & 30.89 & .8941 & {\textbf{.0884}} & {\textbf{29.93}} & {\textbf{.8643}} & {\textbf{.1922}} \\
			Linear Interpolation & {\textbf{30.94}} & {\textbf{.8946}} & .0916 & 29.67 & .8620 & .1982 \\
			\specialrule{0.1em}{1pt}{1pt}
		\end{tabular}
	}
	\vspace{-0.8em}
	\caption{{\bf{Ablation studies on the effect of trajectory representations.}} The results demonstrate that spline-based methods perform better than that of directly optimizing $N$ poses. It also demonstrates that linear interpolation achieves comparably performance as that of cubic B-Spline, due to the short time interval within camera exposure.}
	\label{table_spline}
	\vspace{-1.6em}
\end{table}
\begin{table*}
	\setlength\tabcolsep{2pt}
	\parbox{\textwidth}{
		\resizebox{\linewidth}{!}{
		\begin{tabular}{c|ccc|ccc|ccc|ccc|ccc|ccc}
			\specialrule{0.1em}{1pt}{1pt}
			& \multicolumn{3}{c|}{Cozy2room} & \multicolumn{3}{c|}{Factory} & \multicolumn{3}{c|}{Pool} & \multicolumn{3}{c|}{Tanabata} & \multicolumn{3}{c|}{Trolley} & \multicolumn{3}{c}{Average} \\
			& PSNR$\uparrow$ & SSIM$\uparrow$ & LPIPS$\downarrow$ & PSNR$\uparrow$ & SSIM$\uparrow$ & LPIPS$\downarrow$ & PSNR$\uparrow$ & SSIM$\uparrow$ & LPIPS$\downarrow$ & PSNR$\uparrow$ & SSIM$\uparrow$ & LPIPS$\downarrow$ & PSNR$\uparrow$ & SSIM$\uparrow$ & LPIPS$\downarrow$ & PSNR$\uparrow$ & SSIM$\uparrow$ & LPIPS$\downarrow$ \\
			\specialrule{0.05em}{1pt}{1pt}
			Park \cite{park2017joint} &23.82 &.7221 &.2020 &21.02 &.5090 &.4193 &27.98 &.7258 &.2305 &17.91 &.4637 &.4030 &19.96 &.5610 &.3222 &22.14 &.5963 &.3154\\
			MPR \cite{zamir2021multi} &29.90 &.8862 &.0915 &25.07 &.6994 &.2409 &33.28 &{\textbf{.8938}} &.1290 & 22.60 & .7203 & .2507 & 26.24 & .8356 & .1762 & 27.42 & .8071 & .1777 \\
			PVD \cite{son2021recurrent} &28.06 & .8443 & .1515 & 24.57 & .6877 & .3150 & 30.38 & .8393 & .1977 & 22.54 & .6872 & .3351 & 24.44 & .7746 & .2600 & 26.00 & .7666 & .2519 \\
			SRNDeblur \cite{Tao2018CVPR} &29.47 & .8759 & .0950  & 26.54 & .7604 & .2404 & 32.94 & .8847 & .1045 & 23.20 & .7274 & .2438 & 25.36 & .8119 & .1618 & 27.50 & .8121 & .1691 \\
			DeblurNeRF \cite{deblur-nerf} &25.96 & .7979 & .1024 & 23.21 & .6487 & .2618 & 31.21 & .8518 & .1382 & 22.46 & .6946 & .2455 & 24.94 & .7923 & .1766 & 25.56 & .7571 & .1849 \\
			DeblurNeRF*\cite{deblur-nerf} &30.26 & .8933 & .0791 & 26.40 & .7991 & .2191 & 32.30 & .8755 & .1345 & 24.56 & .7749 & .2166 & 26.24 & .8254 & .1671 & 27.95 & .8336 & .1633 \\
			\specialrule{0.05em}{1pt}{1pt}
			BAD-NeRF (ours) &{\textbf{32.15}} &{\textbf{.9170}} &{\textbf{.0547}} &{\textbf{32.08}} &{\textbf{.9105}} &{\textbf{.1218}} &{\textbf{33.36}} & .8912 &{\textbf{.0802}} &{\textbf{27.88}} &{\textbf{.8642}} &{\textbf{.1179}} &{\textbf{29.25}} &{\textbf{.8892}} &{\textbf{.0833}} &{\textbf{30.94}} &{\textbf{.8946}} &{\textbf{.0916}}\\
			\specialrule{0.1em}{1pt}{1pt}
	\end{tabular}}
    \vspace{-1.0em}
	\caption{{\textbf{Quantitative deblurring comparisons on the synthetic dataset of Deblur-NeRF \cite{deblur-nerf}}.} Note that DeblurNeRF* is trained with the ground-truth poses, while the other one is trained with the estimated poses by COLMAP \cite{colmap}. The experimental results demonstrate that our method achieves the best performance over prior methods. It also demonstrates that the DeblurNeRF is sensitive to the accuracy of the provided camera poses.}
	\label{table_deblur_deblurNeRF}}
    \parbox{\linewidth}{
    \resizebox{\linewidth}{!}{
	\begin{tabular}{c|ccc|ccc|ccc|ccc|ccc|ccc}
		\specialrule{0.1em}{1pt}{1pt}
		& \multicolumn{3}{c|}{Cozy2room} & \multicolumn{3}{c|}{Factory} & \multicolumn{3}{c|}{Pool} & \multicolumn{3}{c|}{Tanabata} & \multicolumn{3}{c|}{Trolley} & \multicolumn{3}{c}{Average} \\
		& PSNR$\uparrow$ & SSIM$\uparrow$ & LPIPS$\downarrow$ & PSNR$\uparrow$ & SSIM$\uparrow$ & LPIPS$\downarrow$ & PSNR$\uparrow$ & SSIM$\uparrow$ & LPIPS$\downarrow$ & PSNR$\uparrow$ & SSIM$\uparrow$ & LPIPS$\downarrow$ & PSNR$\uparrow$ & SSIM$\uparrow$ & LPIPS$\downarrow$ & PSNR$\uparrow$ & SSIM$\uparrow$ & LPIPS$\downarrow$ \\
		\specialrule{0.05em}{1pt}{1pt}
		NeRF+Park & 23.44 & .7024 & .2634 & 20.83 & .5041 & .4133 & 28.69 & .7512 & .2865 & 19.29 & .5317 & .4342 & 20.73 & .6012 & .3804 & 22.60 & .6181 & .3556 \\
		NeRF+MPR & 27.17 & .8334 & .1196 & 23.78 & .6375 & .2499 & 31.15 & .8402 & .1837 & 21.24 & .6914 & .2801 & 26.14 & .8154 & .1979 & 25.90 & .7636 & .2062 \\
		NeRF+PVD & 26.26 & .7977 & .1764 & 23.88 & .6450  & .3074 & 29.02 & .7792 & .2287 & 21.03 & .6566 & .3406 & 23.96 & .7502 & .2772 & 24.83 & .7257 & .2661 \\
		NeRF+SRNDeblur & 27.27 & .8321 & .1261 & 26.19 & .7494 & .2274 & 31.09 & .8375 & .1770 & 21.46 & .6943 & .2839 & 25.01 & .7883 & .2077 & 26.20 & .7803 & .2044 \\
		Deblur-NeRF & 26.05 & .8084 & .1072 & 25.17 & .7253 & .2447 & 30.97 & .8447 & .1554 & 21.77 & .7172 & .2515 & 24.45 & .7785 & .2088 & 25.68 & .7748 & .1935 \\
		Deblur-NeRF* & 29.88 & .8901 & .0747 & 26.06 & .8023 & .2106 & 30.94 & .8399 & .1694 & 22.56 & .7639 & .2285 & 25.78 & .8122 & .1797 & 27.04 & .8217 & .1726 \\
		\specialrule{0.05em}{1pt}{1pt}
		BAD-NeRF (ours) & {\textbf{30.97}} & {\textbf{.9014}} & {\textbf{.0552}} & {\textbf{31.65}} & {\textbf{.9037}} & {\textbf{.1228}} & {\textbf{31.72}} & {\textbf{.8580}} & {\textbf{.1153}} & {\textbf{23.82}} & {\textbf{.8311}} & {\textbf{.1378}} & {\textbf{28.25}} & {\textbf{.8727}} & {\textbf{.0914}} & {\textbf{29.28}} & {\textbf{.8734}} & {\textbf{.1045}}\\
		\specialrule{0.1em}{1pt}{1pt}
	\end{tabular}}
    \vspace{-0.9em}
    \caption{{\textbf{Quantitative novel view synthesis comparisons on the synthetic dataset of Deblur-NeRF \cite{deblur-nerf}.}} The experimental results demonstrate that our method delivers state-of-the-art performance compared to prior works.} 
    \label{table_deblur_archviz}}
	\vspace{-1.7em}
\end{table*}

\PAR{Implementation and training details.}
We implement our method with PyTorch. We adopt the MLP network (\ie $\bF_\theta$) structure of the original NeRF from Mildenhall \etal \cite{nerf} without any modification. Both the network parameters and the camera poses (\ie $\bT_{\mathrm{start}}$ and $\bT_{\mathrm{end}}$) are optimized with the two separate Adam optimizer\cite{kingma2014adam}. The learning rate of the NeRF optimizer and pose optimizer exponentially decays from $5 \times 10^{-4}$ to $5 \times 10^{-5}$ and $1 \times 10^{-3}$ to $1 \times 10^{-5}$. We set number of interpolated poses between $\bT_{\mathrm{start}}$ and $\bT_{\mathrm{end}}$ ($n$ in Eq. \ref{eq_blur_im_formation}) as 7. A total number of 128 points are sampled along each ray. We train our model for 200K iterations on an NVIDIA RTX 3090 GPU. We use COLMAP \cite{colmap} to initialize the camera poses for our method.
%
\subsection{Ablation study}
\vspace{-0.7em}
\PAR{Number of virtual poses.} We evaluate the effect of the number of interpolated virtual cameras (\ie for virtual image synthesis in \eqnref{eq_blur_im_formation}) within exposure time. We choose two sequences from the synthetic dataset of Deblur-NeRF \cite{deblur-nerf} for experiment, \ie the {\textit{cozy2room}} sequence and the {\textit{factory}} sequence, which represent sequences with low-level and high-level motion blur respectively. The experiments are conducted by training our network with a varying number of interpolated virtual cameras. The PSNR metrics computed and plotted in \figref{fig_ablation}. The experimental results demonstrate that the number of virtual cameras does not affect much for images with low-level motion blur. As expected, more virtual cameras are required for images with high-level motion blur. By compromising the image rendering quality and the training complexity (\ie the larger the number of virtual images, the more computational resource is required for training), we choose 7 virtual images for our experiments.
%

\PAR{Trajectory representations.} To evaluate the effect of different trajectory representations, we conduct three experiments: the first is based on optimizing $N$ (\ie $N=7$) camera poses directly, the second is based on optimizing $\bT_\mathrm{start}$ and $\bT_\mathrm{end}$ to represent a linear trajectory, and the last is based on a higher order spline (i.e. cubic B-Spline) which jointly optimizes 4 control knots $\textbf{T}_\mathrm{1}$, $\textbf{T}_\mathrm{2}$, $\textbf{T}_\mathrm{3}$ and $\textbf{T}_\mathrm{4}$ to represent more complex camera motions. For more detailed formulation of cubic B-Spline, please refer to our supplementary material. Since directly optimizing poses would lose the ordering information, we compute the metrics (\eg PSNR) for all 7 poses and choose the best one for comparison. The average quantitative results on the datasets of \textit{Deblur-NeRF} \cite{deblur-nerf} (Cozy2room, Factory, Pool, Tanabata and Trolley) and \textit{MBA-VO} \cite{mba-vo} (ArchViz-low and ArchViz-high) are shown in Table \ref{table_spline}. It demonstrates that directly optimizing $N$ camera poses performs worse than spline-based methods. It also demonstrates that linear interpolation performs comparably as that of cubic B-Spline interpolation. In particular, linear interpolation achieves slightly better performance on the datasets of \textit{Deblur-NeRF}, and cubic B-Spline performs slightly better on that of \textit{MBA-VO}, compared to linear motion model. It is due to the relatively short time interval for camera exposure. Linear interpolation is already sufficient to represent the camera motion trajectory accurately within such short time interval.

%
\begin{table}
	\setlength\tabcolsep{1pt}
	\resizebox{\linewidth}{!}{
		\begin{tabular}{c|ccc|ccc}
			\specialrule{0.1em}{1pt}{1pt}
			& \multicolumn{3}{c|}{ArchViz-low} & \multicolumn{3}{c}{ArchViz-high} \\
			& PSNR$\uparrow$ & SSIM$\uparrow$ & LPIPS$\downarrow$ & PSNR$\uparrow$ & SSIM$\uparrow$ & LPIPS$\downarrow$ \\
			\specialrule{0.05em}{1pt}{1pt}
			Park \cite{park2017joint} & 21.08 & .6032 & .3524 & 21.10 & .5963 & .4243 \\
			MPR \cite{zamir2021multi} & 29.60 & .8757 & .2103 & 25.04 & .7711 & .3576 \\
			PVD \cite{son2021recurrent} & 27.82 & .8318 & .2475 & 25.59 & .7792 & .3441 \\
			SRNDeblur \cite{Tao2018CVPR} & 30.15 & .8814 & .1703 & 27.07 & .8190 & .2796 \\
			Deblur-NeRF \cite{deblur-nerf} & 28.06 & .8491 & .2036 & 25.76 & .7832 & .3277 \\
			Deblur-NeRF* \cite{deblur-nerf} & 29.65 & .8744 & .1764 & 26.44 & .8010 & .3172 \\
			\specialrule{0.05em}{1pt}{1pt}
			BAD-NeRF (ours) & {\textbf{31.27}} & {\textbf{.9005}} & {\textbf{.1503}} & {\textbf{28.07}} & {\textbf{.8234}} & {\textbf{.2460}} \\
			\specialrule{0.1em}{1pt}{1pt}
	\end{tabular}}
	\vspace{-0.7em}
	\caption{{\textbf{Quantitative deblurring comparisons on the synthetic dataset of MBA-VO \cite{mba-vo}}.} The experimental results demonstrate that our method achieves the best performance even with a camera that is not moving at a constant velocity within exposure time.}
	\label{table_archviz}
	\vspace{-1.8em}
\end{table}
\begin{figure*}
	\setlength\tabcolsep{1pt}
	\centering
	\begin{tabular}{cccccccc}
		&\raisebox{-0.02in}{\rotatebox[origin=t]{90}{\scriptsize Input}} &
		\includegraphics[valign=m,width=0.89\textwidth]{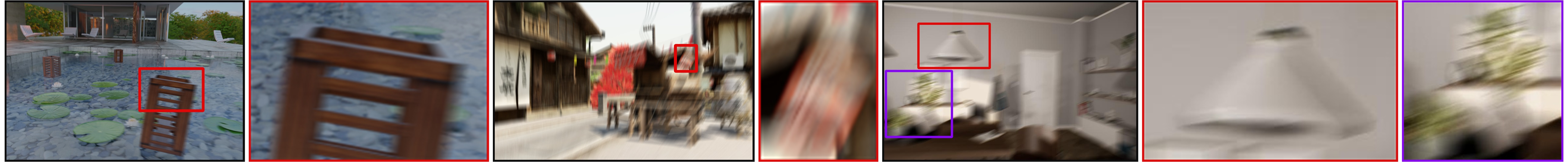}\\
		\specialrule{0em}{.05em}{.05em}
		&\raisebox{-0.035in}{\rotatebox[origin=t]{90}{\scriptsize MPR \cite{zamir2021multi}}} &
		\includegraphics[valign=m,width=0.89\textwidth]{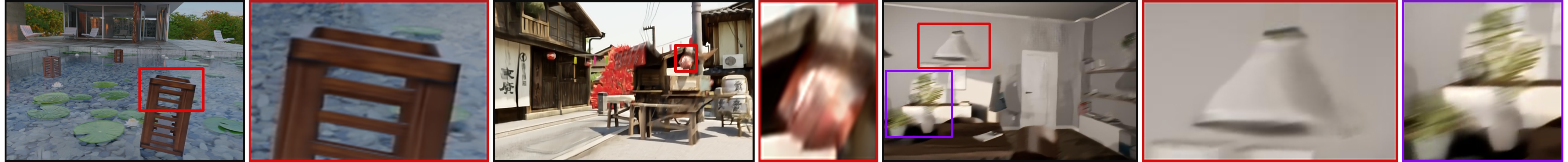}\\
		\specialrule{0em}{.05em}{.05em}
		&\raisebox{-0.035in}{\rotatebox[origin=t]{90}{\scriptsize PVD \cite{son2021recurrent}}} &
		\includegraphics[valign=m,width=0.89\textwidth]{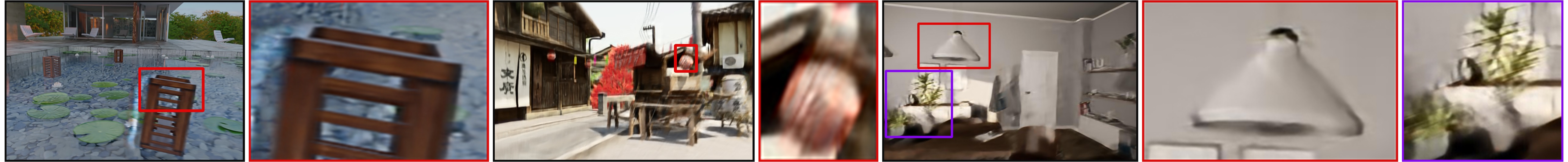}\\
		\specialrule{0em}{.05em}{.05em}
		&\raisebox{-0.035in}{\rotatebox[origin=t]{90}{\scriptsize SRN \cite{Tao2018CVPR}}} &
		\includegraphics[valign=m,width=0.89\textwidth]{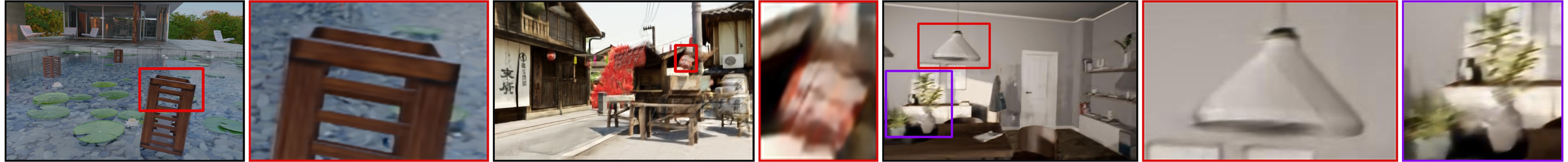}\\
		\specialrule{0em}{.05em}{.05em}
		\raisebox{-0.035in}{\rotatebox[origin=t]{90}{\scriptsize Deblur-}}
		&\raisebox{-0.035in}{\rotatebox[origin=t]{90}{\scriptsize NeRF* \cite{deblur-nerf}}} &
		\includegraphics[valign=m,width=0.89\textwidth]{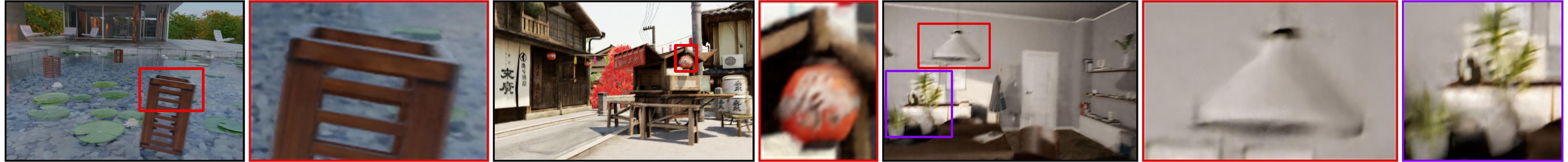}\\
		\specialrule{0em}{.05em}{.05em}
		&\raisebox{-0.035in}{\rotatebox[origin=t]{90}{\scriptsize BAD-NeRF}} &
		\includegraphics[valign=m,width=0.89\textwidth]{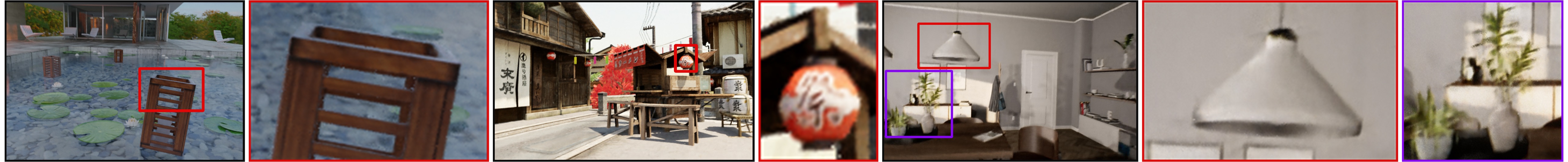}\\
		\specialrule{0em}{.05em}{.05em}
		&\raisebox{-0.035in}{\rotatebox[origin=t]{90}{\scriptsize Ground-truth}} &
		\includegraphics[valign=m,width=0.89\textwidth]{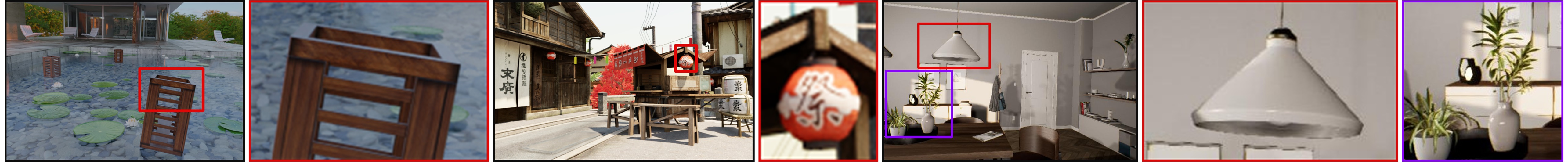}\\
	\end{tabular}
	\vspace{-0.5em}
	\caption{ {\textbf{Qualitative results of different methods with synthetic datasets.}} BAD-NeRF achieves the best performance under inaccurate poses on various scenes and different levels of blur. Since DeblurNeRF does not explicitly model occlusions, it fails to render sharp edges (\ie result in the second column) where large depth change exists.}
	\label{fig:comparison-tencent-syn}
	\vspace{-1em}
\end{figure*}
\vspace{-0.05em}
\subsection{Results}
\vspace{-0.8em}
\PAR{Quantitative evaluation results.} For the subsequent experimental results, we use linear interpolation by default, unless explicitly stated. We evaluate the deblurring and novel view image synthesis performance with both the synthetic images from Deblur-NeRF \cite{deblur-nerf} and MBA-VO \cite{mba-vo}. We also evaluate the accuracy of the refined camera poses of our method against that from Park \cite{park2017joint} and BARF \cite{Lin2021} in terms of the ATE metric. Both \tabnref{table_deblur_deblurNeRF} and \tabnref{table_deblur_archviz} present the experimental results in deblurring and novel view image synthesis respectively, with the dataset from Deblur-NeRF \cite{deblur-nerf}. It reveals that single image based deblurring methods, \eg MPR \cite{zamir2021multi}, PVD \cite{son2021recurrent} and SRNDeblur \cite{Tao2018CVPR} fail to outperform our method, due to the limited information possessed by a single blurry image and the deblurring network struggles to restore the sharp image. The results also reveal that classic multi-view image deblurring method, i.e. the work from Park et al.\cite{park2017joint}, cannot outperform our method, thanks to the powerful representation capability of deep neural networks for our method. In the experiments, we also investigate the effect of the input camera poses accuracy to DeblurNeRF \cite{deblur-nerf}. We conduct two experiments, \ie the DeblurNeRF network is trained with ground truth poses and with that computed by COLMAP \cite{colmap}. Since the images are motion blurred, the poses estimated by COLMAP are not accurate. The results reveal that the DeblurNeRF network trained with poses from COLMAP \cite{colmap} performs poorly compared to that of ground truth poses. It further demonstrates that DeblurNeRF is sensitive to the accuracy of the camera poses, since they do not optimize them during training.

To better evaluate the performance of our network, we also conduct experiments with the dataset from MBA-VO \cite{mba-vo}. The images from MBA-VO \cite{mba-vo} are generated by using real camera motion trajectories from ETH3D dataset \cite{schops2019bad}. The motion is not at constant velocity compared to that of the dataset from DeblurNeRF \cite{deblur-nerf}. The experimental results presented in \tabnref{table_archviz} demonstrate that our network also outperforms other methods, even with a camera that is not moving in constant velocity within exposure time. 
\begin{table}
	\setlength\tabcolsep{1pt}
	\centering
	\resizebox{\linewidth}{!}{
		\begin{tabular}{c|cccccc}
			\specialrule{0.1em}{1pt}{1pt}
				& \footnotesize Cozy2room	& \footnotesize Factory	& \footnotesize Pool
				& \footnotesize Tanabata	& \footnotesize Trolley\\
			\hline
			\scriptsize {COLMAP-blur\cite{colmap}}
				& \footnotesize.128$\pm$.090 & \footnotesize.148$\pm$.093 & \footnotesize.057$\pm$.026
				& \footnotesize.103$\pm$.090 & \footnotesize.045$\pm$.042\\
			\scriptsize {BARF\cite{Lin2021}}
				& \footnotesize.291$\pm$.111 & \footnotesize.145$\pm$.088 & \footnotesize.083$\pm$.036
				& \footnotesize.203$\pm$.091 & \footnotesize.244$\pm$.074\\
			\scriptsize {BAD-NeRF (ours)}
				& \footnotesize.050$\pm$.025 & \footnotesize.033$\pm$.012 & \footnotesize.020$\pm$.007
				& \footnotesize.016$\pm$.008 & \footnotesize.007$\pm$.004\\
			\specialrule{0.1em}{1pt}{1pt}
		\end{tabular}
	}
	\vspace{-0.8em}
	\caption{{\textbf{Pose estimation performance of BAD-NeRF on various blur sequences.}} The results are in the absolute trajectory error metric (ATE). The COLMAP-blur represents the result of COLMAP with blurry images.}
	\label{table_ate}
	\vspace{-2em}
\end{table}
\begin{figure*}[!ht]
	\setlength\tabcolsep{1pt}
	\centering
	\begin{tabular}{cccccc}
		\includegraphics[width=0.145\textwidth]{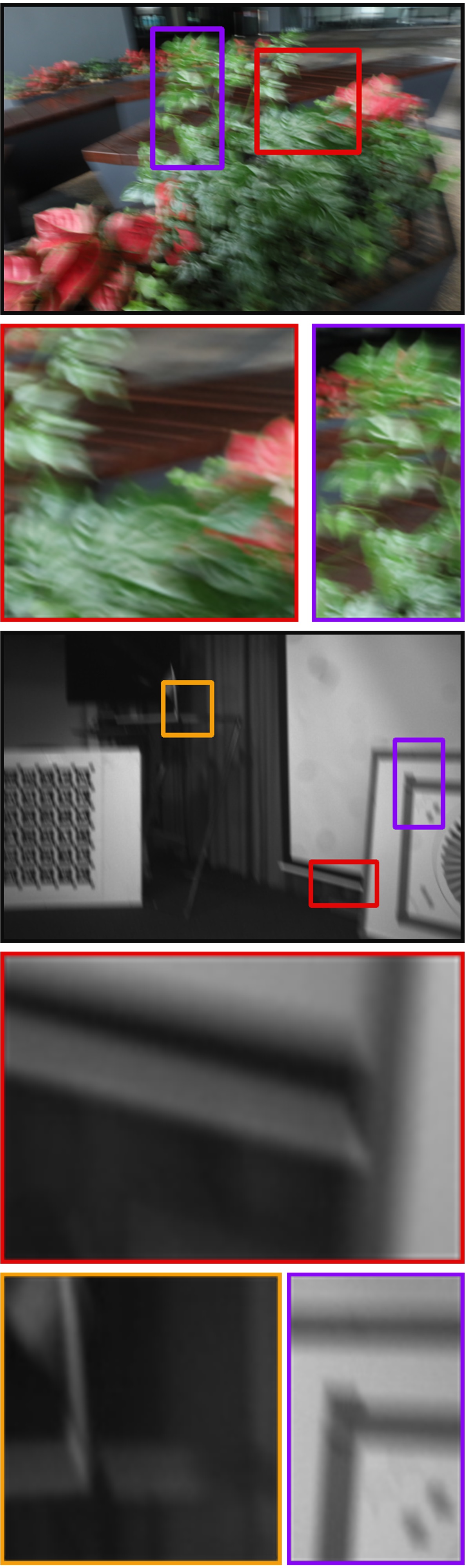} &
		\includegraphics[width=0.145\textwidth]{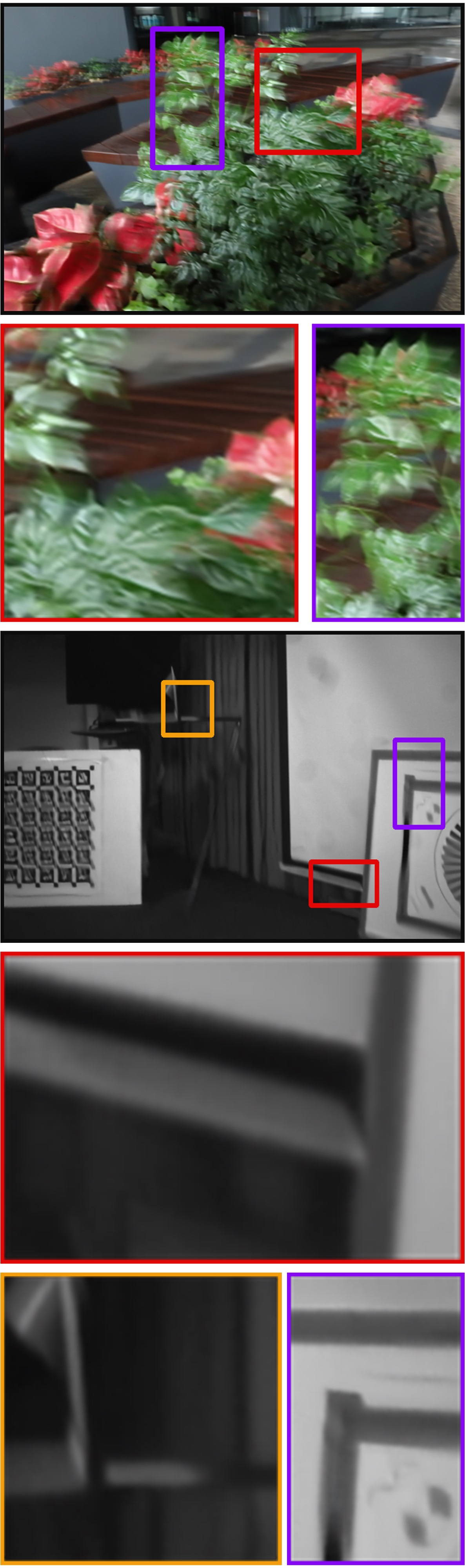} &
		\includegraphics[width=0.145\textwidth]{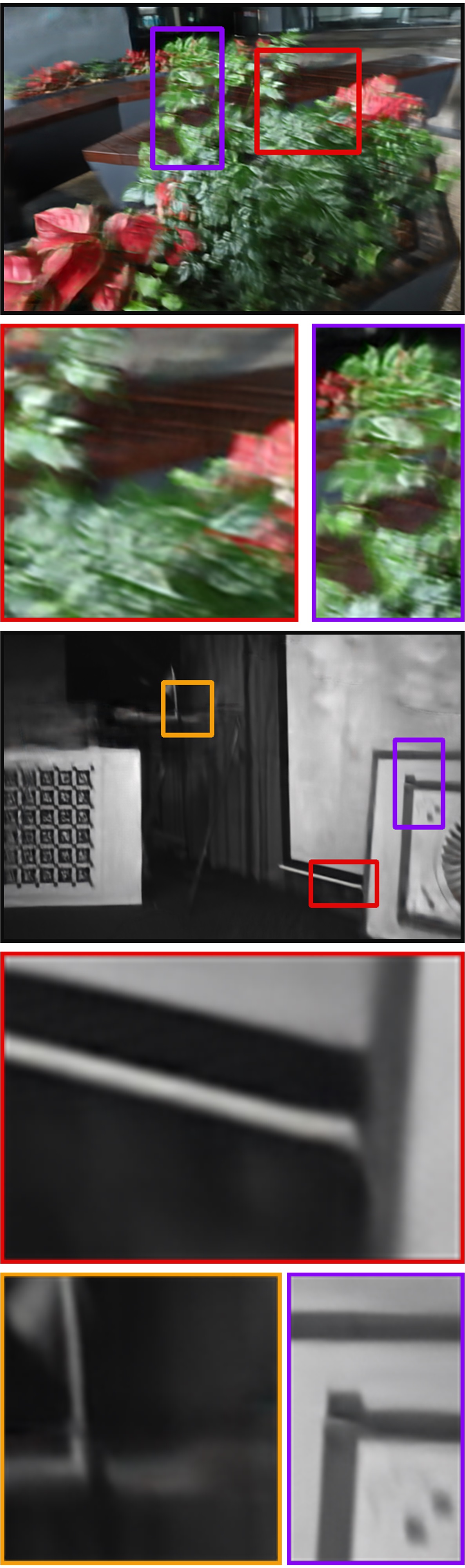} &
		\includegraphics[width=0.145\textwidth]{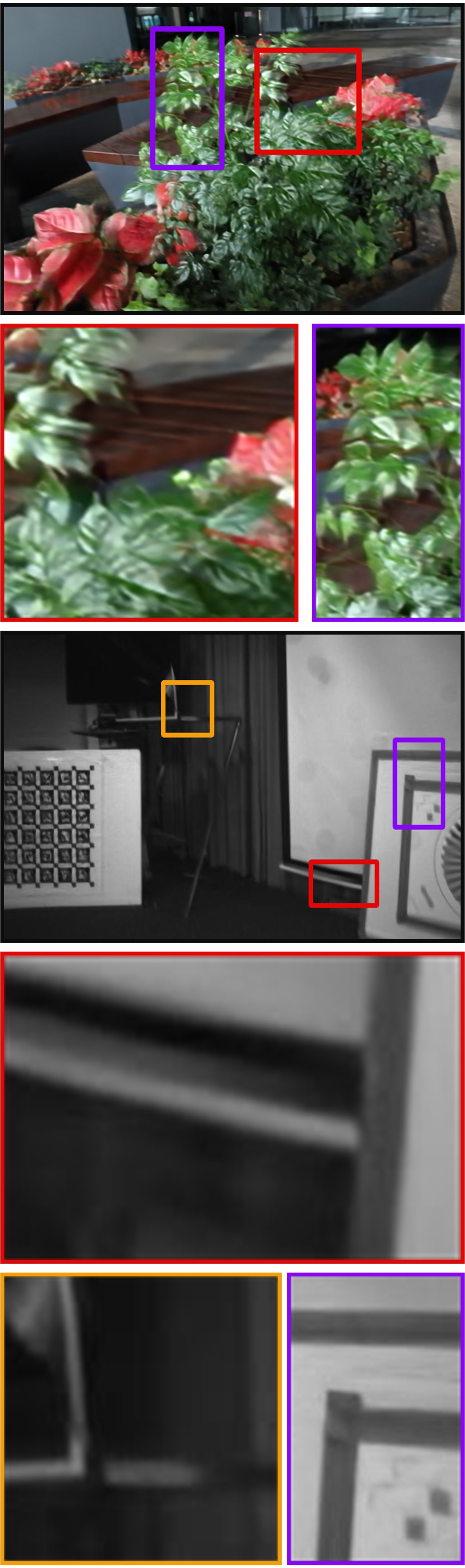} &
		\includegraphics[width=0.145\textwidth]{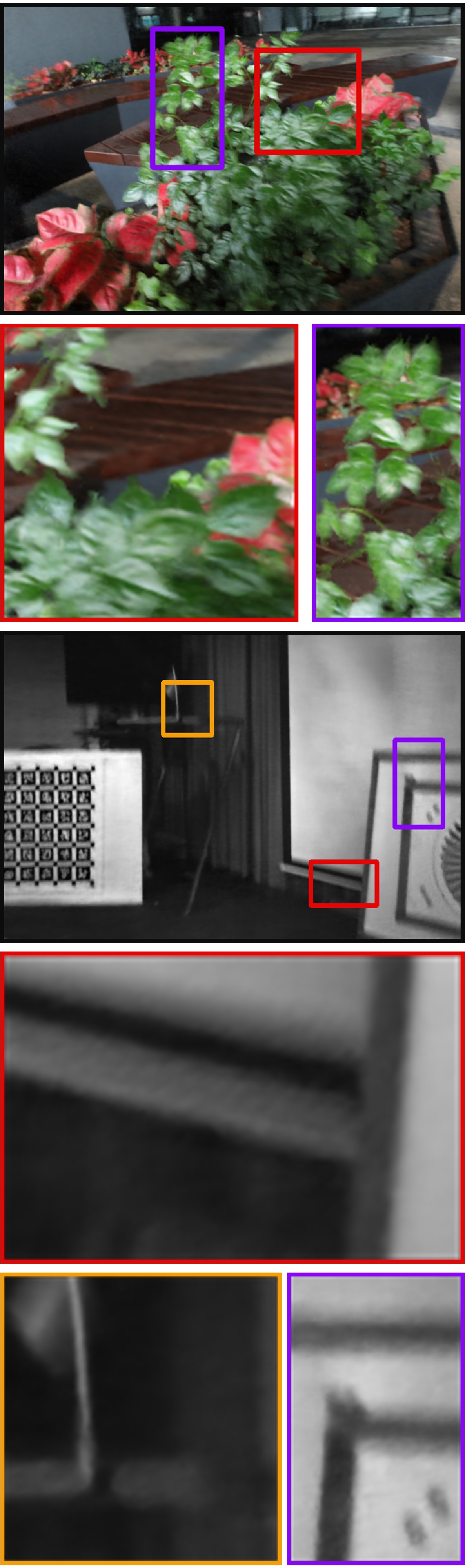} &
		\includegraphics[width=0.145\textwidth]{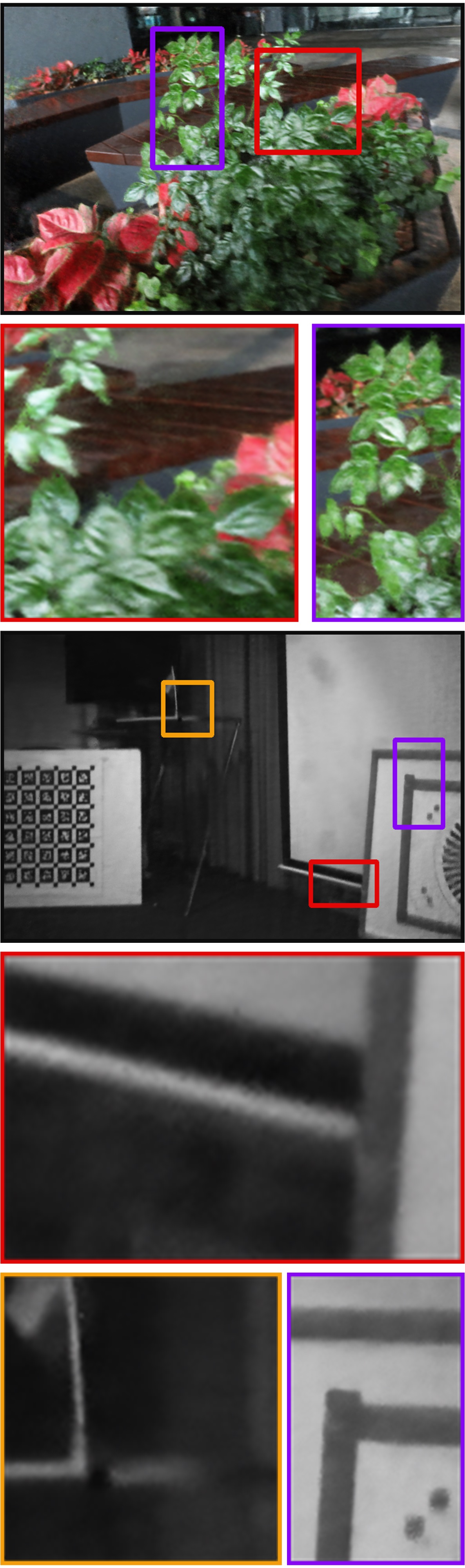} \\
		\specialrule{0em}{0.05pt}{0.05pt}
		Input & MPR \cite{zamir2021multi} & PVD \cite{son2021recurrent} & SRN \cite{Tao2018CVPR} & \small{Deblur-NeRF} \cite{deblur-nerf} & BAD-NeRF
	\end{tabular}
	\vspace{-0.7em}
	\caption{{\textbf{Qualitative results of different methods with the real datasets.}} The experimental results demonstrate that our method achieves superior performance over prior methods on the real dataset as well. Best viewed in high resolution.}
	\label{fig:comparison-tencent-real}
	\vspace{-1.5em}
\end{figure*}
To evaluate the performance of camera pose estimation, we also tried to compare our method against the work from Park \etal \cite{park2017joint}. However, we found that the method from Park \etal \cite{park2017joint} hardly converges and we did not list their metrics. We therefore only present the comparisons of our method against that of COLMAP \cite{colmap} and BARF \cite{Lin2021}. The experiments are conducted on the datasets from DeblurNeRF \cite{deblur-nerf}. The estimated camera poses are aligned with the ground truth poses before the absolute trajectory error metric is computed. The experimental results presented in \tabnref{table_ate} demonstrate that our method can recover the camera poses more accurately.

\PAR{Qualitative evaluation results.} 
We also evaluate the qualitative performance of our method against the other methods. The experiments are conducted on both synthetic and real-world datasets. The experimental results presented in both \figref{fig:comparison-tencent-syn} and \figref{fig:comparison-tencent-real} demonstrate that our method also outperforms other methods as in the quantitative evaluation section. In particular, single image deblurring networks (\ie MPR \cite{zamir2021multi}, PVD \cite{son2021recurrent} and SRNDeblur \cite{Tao2018CVPR}) indeed can achieve impressive performance on some images, which are not severely blurred. However, they fail to restore the sharp image and bring in unpleasing artifacts for severely blurred images. On the contrary, our method always delivers consistent performance regardless of the level of motion blur. The reason is that our method takes advantage of multi-view images to recover a consistent 3D representation of the scene. It learns to fuse information from other views to improve the deblurring performance. 
Our method also delivers better results compared to DeblurNeRF \cite{deblur-nerf}. One reason is that DeblurNeRF does not optimize the camera poses/motion trajectories within exposure time. They can deliver impressive results if the accurate poses are known. However, the performance would degrade if inaccurate poses are provided. Unfortunately, it is usually not trivial to recover accurate camera poses from motion blurred images, especially when the images are severely blurred. Thanks to the explicit modeling of the camera motion trajectory within exposure time, our method does not have such limitations. As shown in \tabnref{table_ate}, our method can accurately estimate the camera poses together with the learning of the network. Another reason is caused by the formulation of DeblurNeRF. The motion blur aware image formation model of DeblurNeRF does not model occlusions. They synthesize a blurry image by convolving the rendered image with a learned point spread function. It thus cannot accurately model the image formation process for pixels/3D points lying around an edge, which has a large change of depth. In contrast, our method follows the real physical image formation process of a motion blurred image and does not have such an issue. \figref{fig:comparison-tencent-syn} clearly shows that DeblurNeRF fails to render sharp images around the edge of the floating box even trained with ground truth poses, caused by the occlusion problem.

\vspace{-0.5em}
\section{Conclusion}
In this paper, we propose a photometric bundle adjustment formulation for motion blurred images by using NeRF to implicitly represent the 3D scene. Our method jointly learns the 3D representation and optimizes the camera poses with blurry images and inaccurate initial poses. Extensive experimental evaluations with both real and synthetic datasets are conducted. The experimental results demonstrate that our method can effectively deblur images, render novel view images and recover the camera motion trajectories accurately within exposure time. 

\PAR{Acknowledgements.} This work was supported in part by NSFC under Grant 62202389, in part by a grant from the Westlake University-Muyuan Joint Research Institute, and in part by the Westlake Education Foundation. 

{\small
	\bibliographystyle{ieee_fullname}
	\bibliography{egbib,bibography_peidong}
}

\end{document}